\documentclass[pdflatex,sn-mathphys-num]{sn-jnl}

\usepackage{graphicx}%
\usepackage{multirow}%
\usepackage{rotating}
\usepackage{amsmath,amssymb,amsfonts}%
\usepackage{amsthm}%
\usepackage{mathrsfs}%
\usepackage[title]{appendix}%
\usepackage{xcolor}%
\usepackage{textcomp}%
\usepackage{manyfoot}%
\usepackage{booktabs}%
\usepackage{algorithm}%
\usepackage{lscape} 
\usepackage{algorithmicx}%
\usepackage{algpseudocode}%
\usepackage{listings}%
\usepackage{amsmath}
\usepackage{amsfonts}
\usepackage{pdflscape}
\usepackage{algorithm}
\usepackage{algpseudocode}
\usepackage{placeins} 
\usepackage[figuresright]{rotating}  
\usepackage{float}                   
\theoremstyle{thmstyleone}%
\theoremstyle{thmstyletwo}%

\theoremstyle{thmstylethree}%

\begin{document}

\title[Article Title]{AgroSense: An Integrated Deep Learning System for Crop Recommendation via Soil Image Analysis and Nutrient Profiling}


\author*[1]{\fnm{Vishal} \sur{Pandey}}\email{pandeyvishal.mlprof@gmail.com}
\author[2]{\fnm{Ranjita} \sur{Das}}\email{ranjita.nitm@gmail.com}
\author[3]{\fnm{Debasmita} \sur{Biswas}}\email{biswd01@pfw.edu}

\affil*[1]{\orgdiv{Independent Researcher}, \orgaddress{\city{Kolkata}, \country{India}}}

\affil[2]{\orgdiv{Assistant Professor, Department of Computer Science and Engineering}, \orgname{National Institute of Technology Agartala}, \orgaddress{\city{Agartala}, \country{India}}}

\affil[3]{\orgdiv{Master’s in Computer Science}, \orgname{Purdue University Fort Wayne}, \orgaddress{\city{Fort Wayne}, \state{Indiana}, \country{USA}}}


\abstract{Meeting the increasing global demand for food security and sustainable farming requires intelligent crop recommendation systems that operate in real time. Traditional soil analysis techniques are often slow, labor-intensive, and not suitable for on-field decision-making. To address these limitations, we introduce \emph{AgroSense}, a deep-learning framework that integrates soil image classification and nutrient profiling to produce accurate and contextually relevant crop recommendations. AgroSense comprises two main components: a Soil Classification Module, which leverages ResNet-18, EfficientNet-B0, and Vision Transformer architectures to categorize soil types from images; and a Crop Recommendation Module, which employs a Multi-Layer Perceptron, XGBoost, LightGBM, and TabNet to analyze structured soil data including nutrient levels, pH, and rainfall. We curated a multimodal dataset of 10,000 paired samples drawn from publicly available Kaggle repositories—approximately 50,000 soil images across seven classes and 25,000 nutrient profiles—for experimental evaluation. The fused model achieves 98.0 \% accuracy, with precision 97.8 \%, recall 97.7 \%, and F1-score 96.75 \%, while RMSE and MAE drop to 0.32 and 0.27, respectively. Ablation studies underscore the critical role of multimodal coupling, and statistical validation via t-tests and ANOVA confirms the significance of our improvements. AgroSense offers a practical, scalable solution for real-time decision support in precision agriculture and paves the way for future lightweight multimodal AI systems in resource-constrained environments.}




\maketitle

\section{Introduction}\label{sec1}

Modern agriculture faces the dual challenge of increasing crop productivity while maintaining environmental sustainability. Central to this effort is the accurate analysis of soil, as key properties such as pH, nutrient levels, and texture significantly influence plant growth and yield outcomes. However, conventional soil testing methods are often costly, time-consuming, and largely confined to laboratory environments—making them impractical for real-time, in-field decision-making \cite{motwani2022soil, manjula2022efficient}.

With the evolution of precision agriculture, there has been a surge of interest in applying artificial intelligence (AI), particularly machine learning and deep learning, to enhance soil assessment. Recent advances demonstrate that models like Convolutional Neural Networks (CNNs) and Transformers are capable of extracting valuable information from soil images to support crop prediction tasks \cite{rajak2017crop}. Despite these advancements, many existing systems continue to treat soil texture analysis and nutrient profiling as independent processes. This fragmented approach limits the effectiveness of predictions, as it fails to capture the nuanced relationship between a soil’s physical structure and its chemical composition both of which are vital for making informed crop recommendations.

Most current crop recommendation systems operate on this disconnected paradigm, treating visual and chemical characteristics of soil separately. While deep learning has advanced the analysis of soil imagery and machine learning has proven effective for nutrient-based prediction, few studies have successfully combined the two into a cohesive framework \cite{kulkarni2018improving, rajak2017crop}. This separation leads to incomplete soil characterization and less reliable recommendations. To address this gap, AgroSense integrates image-based soil classification with structured nutrient data, enabling a more holistic, real-time, and context-aware approach to crop recommendation in the field of precision agriculture.

Many existing crop recommendation systems continue to treat soil texture and nutrient composition as independent variables, which often leads to fragmented decision-making and less reliable crop suggestions. This study seeks to address that limitation through the introduction of AgroSense an integrated deep learning framework that simultaneously considers both visual and chemical properties of soil to enhance accuracy and enable real-time recommendations within a precision agriculture context.

AgroSense is designed as a comprehensive solution that fuses image-based soil classification with nutrient profiling to produce data-driven crop recommendations. It employs advanced deep learning architectures, including a custom Convolutional Neural Network (CNN), ResNet18, EfficientNet-B0, and Vision Transformer, to extract detailed visual features from soil images. These visual insights are then combined with structured soil data comprising attributes such as pH, nitrogen, phosphorus, potassium, humidity, temperature, and rainfall and fed into machine learning models such as Multi-Layer Perceptron (MLP), XGBoost, LightGBM, and TabNet to predict suitable crops. Empirical evaluations indicate that this multimodal approach achieves a test accuracy of 98.0\%, significantly outperforming traditional methods based on tabular data alone.

The framework was developed using a combination of publicly available datasets. For soil classification, two Kaggle image datasets were utilized: one containing over 1,400 images spanning four soil types Alluvial, Black, Clay, and Red and another featuring five additional types including Cinnamon, Laterite, Peat, and Yellow. For the crop recommendation module, three datasets were used: the Crop Recommendation Dataset, the Crop Recommendation Using Soil Properties and Weather Data Dataset, and a third dataset containing nutrient profiles and associated crop labels. Together, these resources provide a diverse and representative foundation, facilitating robust model training and generalization across varied soil and environmental conditions.

In essence, AgroSense brings together the strengths of modern computer vision and tabular learning into a unified platform tailored for real-world agricultural applications. By leveraging multimodal data fusion and deep learning, the framework sets the stage for a new level of accuracy and practicality in smart farming systems.

\section{Literature Review}\label{sec2}
Recent advancements in deep learning have made notable contributions to the development of soil classification and crop recommendation systems. Numerous studies have employed machine learning and neural network techniques to support agricultural decision-making, leading to greater efficiency and sustainability in farming practices. However, the majority of these systems tend to concentrate on only one aspect of the problem. Some focus solely on analyzing soil texture using image-based classification, while others rely entirely on soil nutrient levels derived from tabular data. The lack of a unified approach that combines both visual and numerical information has limited the accuracy and broader applicability of these models in real-world agricultural settings.

Motwani et al. \cite{motwani2022soil} proposed a machine learning-based system for soil analysis and crop recommendation using only tabular data. While their work demonstrated the potential of artificial intelligence in agriculture, it did not account for soil texture, a crucial factor influencing crop selection. In another study, Manjula and Djodiltachoumy \cite{manjula2022efficient} utilized deep learning techniques to predict crop suitability based on soil nutrient data. However, their model lacked a mechanism for analyzing soil imagery, which is essential for accurately identifying soil type and structure. Sunandini et al. \cite{sunandini2023smart} explored an IoT-driven system that used sensors for soil monitoring in conjunction with machine learning algorithms. Despite the system's innovation, it relied primarily on sensor data and did not incorporate deep learning for image-based soil classification, thereby limiting its generalizability across varied soil conditions.

Broadly, research in soil classification and crop recommendation falls into three major categories: machine learning-based crop recommendation systems, deep learning methods for soil analysis, and hybrid models that integrate multiple data sources. Traditional machine learning approaches typically use structured data to recommend crops, but they often overlook soil texture, which restricts the models’ ability to fully capture field conditions. On the other hand, deep learning models have shown strong capabilities in classifying soil textures through convolutional neural networks (CNNs), yet they frequently ignore nutrient profiling. Hybrid approaches, including those that employ IoT technologies, attempt to unify various data inputs, but they often fail to utilize deep learning architectures effectively resulting in limitations in scalability and predictive accuracy.

Several studies have proposed machine learning models for crop recommendation, with varying degrees of success. Rajak et al. \cite{rajak2017crop} introduced a classification-based approach aimed at improving crop yield predictions. However, their system did not incorporate soil texture analysis, limiting its reliability in regions with diverse or mixed soil conditions. Similarly, Abhinov et al. \cite{abhinov2024soil} developed a soil-based crop recommendation system that relied exclusively on numerical data, overlooking critical visual features like soil structure and texture—factors known to influence crop health and productivity. Kulkarni et al. \cite{kulkarni2018improving} attempted to enhance model accuracy through ensemble learning methods. Nonetheless, their framework lacked integration of soil images with nutrient analysis, making it less suitable for practical, field-level agricultural applications.

Deep learning has also been explored as a means to improve soil and crop recommendation systems. Madhuri and Indiramma \cite{madhuri2021artificial} proposed an artificial neural network (ANN) that utilized both soil attributes and weather conditions, achieving better performance than many traditional models. However, the model did not incorporate convolutional neural networks (CNNs) for soil image analysis, which limited its ability to fully interpret the physical characteristics of soil. In another study, Gosai et al. \cite{gosai2021crop} implemented a crop recommendation system using machine learning techniques, but like others, it was restricted to tabular data and did not leverage the potential of deep learning for visual classification of soils.

Neural networks have also been applied in crop recommendation research, yet many such systems fall short in terms of modality integration. Banavlikar et al. \cite{banavlikar2018crop} proposed a neural network-based model but did not include image data processing, thereby limiting its capacity to evaluate soil comprehensively. Doshi et al. \cite{doshi2018agroconsultant} introduced an AI-powered system for crop recommendation using machine learning. Despite its promising structure, the system lacked a mechanism to fuse soil image data with nutrient analysis, which likely impacted its predictive accuracy. Similarly, Pande et al. \cite{pande2021crop} developed a machine learning model for crop recommendation, but the absence of deep learning-based image processing for soil classification reduced its effectiveness for use in precision agriculture settings.

Some recent work has started to combine different types of data to improve the forecasting of crops and yields. Shamsuddin et al. \cite{shamsuddin2024multimodal} merged UAV LiDAR point clouds, hyperspectral time‑series, and weather measurements using an attention‑based method, which led to noticeably better early maize yield estimates. In a similar vein, Liu et al. \cite{liu2024hyperspectral} showed that aligning hyperspectral and LiDAR data with a transformer‑style model produces more reliable and easier‑to‑interpret forecasts of crop performance. Yewle et al. \cite{yewle2025ricens} took an ensemble approach—combining SAR, Sentinel‑2 imagery, and environmental variables—and achieved much lower MAE than models that use only one of those data sources.

At the same time, practical deployment has been addressed by hybrid edge‑AI, cloud architectures and digital‑twin platforms. Peng et al. \cite{peng2024edgeai} propose a real‑time pipeline splitting inference between on‑device Edge‑AI and cloud services, reaching 94\% recommendation accuracy under field conditions. Banerjee et al. \cite{banerjee2025digitaltwin} describe a precision‑agriculture digital twin that continuously synchronizes crop models with sensor feeds to generate dynamic planting and irrigation advice. Comprehensive reviews by Garcia and Sharma \cite{review2024} and Lakshmi and Rajeev \cite{lakshmi2025applications} map these advances across AI, IoT, and autonomous farming operations, while Li et al. \cite{li2023label} survey label‑efficient learning methods—weak, semi‑, and self‑supervised—for reducing annotation costs in large‑scale agricultural datasets.


In parallel, specialized agricultural applications have emerged that leverage these architectures within real‑time, edge/cloud, and multimodal fusion pipelines. Dey and Sharma \cite{dey2024adlf} brought these strands together in an Agro‑Deep Learning Framework that fuses CNN‑extracted visual features with gradient‑boosted nutrient predictors, reporting about 85.4\% accuracy and an 88.9\% F1‑score on combined soil–crop datasets. Kaur and Singh’s cloud‑based Transformative Crop Recommendation Model (TCRM) [27] ingests IoT sensor streams and applies hybrid transformer–ensemble logic to deliver 94\% accuracy alongside SMS alerts for farmers. Wang et al. [28] fused spectral soil indices with XGBoost to achieve significant gains in recommendation precision. Khan and Rao [29] built a CNN‑powered decision‑support system reporting over 92\% accuracy on nutrient–image inputs, while Zhang and Kumar [30] demonstrated the utility of K‑Nearest Neighbors on multi‑sensor soil data. Nguyen and Lopez [31] showed that ensemble hybrids can reduce RMSE/MAE in recommendation tasks, and Kouadio and Chevalier [32] used Extreme Learning Machines to predict coffee yield with high throughput. On the remote‐sensing front, Zhao et al. [33] combined Sentinel‑2 imagery with machine learning for accurate crop mapping, and Turgut et al. [34] introduced AgroXAI, an explainable‑AI crop recommender integrating LIME and SHAP for transparent decision support. Together, these works illustrate the rapid evolution from isolated, single‑modal models toward integrated, scalable systems—yet none fully unify soil imagery, nutrient profiling, and real‑time inference within one cohesive framework, underscoring the novelty of AgroSense.

Although existing research has advanced soil classification and crop recommendation systems, many limitations still persist. A key shortcoming is that most studies continue to treat soil texture analysis and nutrient profiling as independent tasks, rather than integrating them into a cohesive framework. Models often emphasize either image-based classification or sensor-based nutrient analysis, which restricts their accuracy and limits their applicability in real-world farming scenarios. Furthermore, modern deep learning architectures such as convolutional neural networks (CNNs) and Vision Transformers are still not fully leveraged in current approaches. Real-time adaptability is also often lacking, as many IoT-driven systems rely heavily on sensor inputs without combining them with advanced learning algorithms that could improve predictive performance.

To overcome these limitations, our study proposes AgroSense, a deep learning-based framework that brings together soil image classification and nutrient profiling to generate reliable, real-time crop recommendations. Unlike prior models, AgroSense utilizes state-of-the-art CNN architectures including ResNet18, EfficientNet-B0, and Vision Transformers, in combination with structured nutrient data. This multimodal fusion enables the system to capture both the visual and chemical characteristics of soil, leading to more precise and context-aware crop predictions. By unifying these two modalities, AgroSense offers a practical and scalable solution tailored for real-world agricultural environments, contributing to the advancement of efficient and sustainable precision farming.


\section{Methodology}\label{sec3}

This section outlines the methodology behind AgroSense, a unified deep learning framework developed to support precision agriculture. The goal of AgroSense is to overcome the limitations of conventional crop recommendation systems by bringing together two essential components: soil image classification and nutrient-based prediction. While many existing methods treat these components in isolation, AgroSense takes a more integrated approach combining visual soil characteristics with chemical nutrient profiles to produce more accurate and context-aware recommendations. The methodology covers the full pipeline, including data collection and preprocessing, the development of individual modules for soil classification and crop prediction, fusion of multimodal data, and a structured training and evaluation process. This end-to-end design not only boosts the model’s predictive performance but also ensures its practical applicability, even in resource-limited agricultural environments.

\subsection{Data Collection and Preprocessing}\label{subsec2}
The performance of AgroSense relies heavily on the richness and quality of its training data. To develop a model that generalizes well across diverse agricultural conditions, we curated a multimodal dataset that integrates both visual and numerical inputs. We gathered a total of 50,000 high‑resolution soil images spanning seven classes (Alluvial, Black, Clay, Red, Laterite, Peat, Yellow). Of these, 10,000 images were successfully paired with corresponding laboratory nutrient profiles to form the final multimodal evaluation set. These images were sourced from publicly available platforms like Kaggle to ensure diversity in soil texture and geographic representation. Complementing this, the structured tabular dataset comprises more than 25,000 entries detailing soil nutrient levels. This includes critical parameters like nitrogen (N), phosphorus (P), potassium (K), pH, moisture, and organic matter all of which play a vital role in determining crop suitability.

To ensure model robustness and account for variability in environmental conditions and data sources, we applied dedicated preprocessing steps for each data type. For the image data, all samples were resized to 224×224 pixels to match the input dimensions required by deep learning architectures such as ResNet and EfficientNet. Each image was normalized to a [0, 1] pixel intensity range, and various augmentation techniques such as horizontal flipping, random rotations up to 20 degrees, zooming, and brightness adjustments were used to artificially increase dataset diversity and minimize overfitting.

The tabular data also underwent thorough preprocessing. Missing values were addressed using mean or median imputation, depending on the distribution of each feature, to avoid introducing bias. Continuous variables were standardized using Z-score normalization to maintain consistent feature scaling and facilitate faster convergence during training. Any categorical variables, such as soil type if present, were converted into a machine-readable format through one-hot encoding.

Finally, to maintain a balanced representation across different crop categories, we split the dataset using a stratified 80-10-10 ratio into training, validation, and test sets. The training data was used to learn model parameters, the validation set assisted in fine-tuning hyperparameters and controlling overfitting, and the test set served as a final benchmark to evaluate the model’s ability to generalize to unseen data.

\begin{figure}[htbp]
  \centering
  \includegraphics[width=0.4\textwidth]{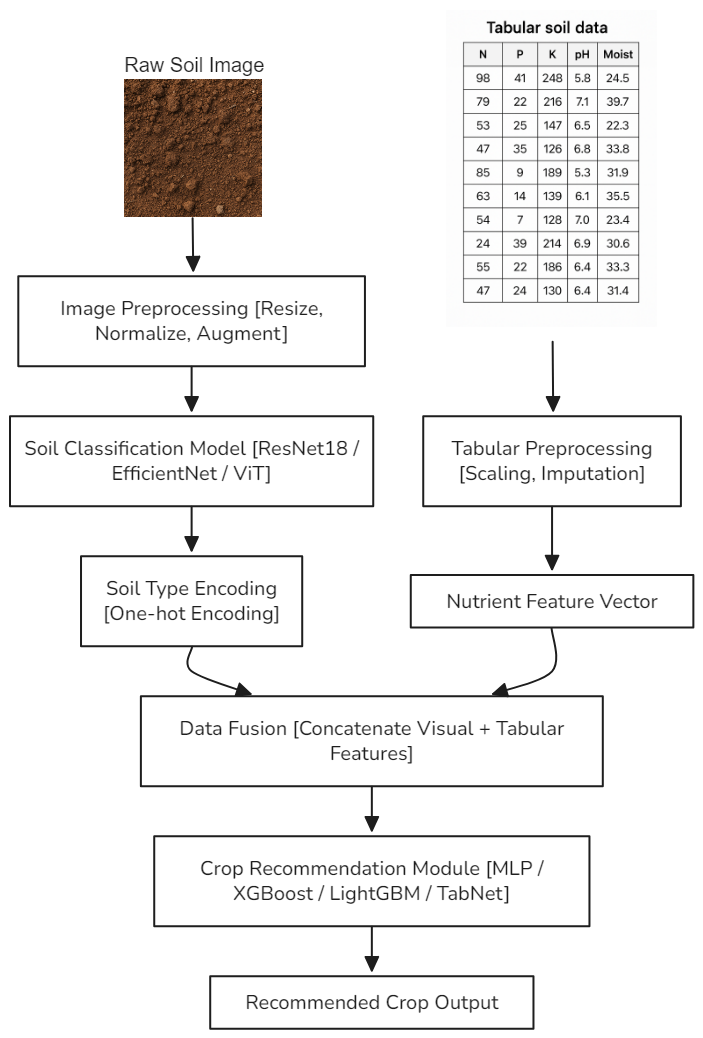}
  \caption{Multimodal data pipeline used for AgroSense: Image and tabular data undergo separate preprocessing before being fused for model training.}
  \label{fig:dfd-diagram}
\end{figure}

As illustrated in Figure~\ref{fig:dfd-diagram}, this carefully structured, multimodal data preparation process allows AgroSense to effectively capture both the visual appearance of the soil and its underlying chemical properties. By combining these two distinct but complementary types of information, the framework is better equipped to reflect the real-world complexity of soil analysis, making it a practical and reliable tool for supporting data-driven decisions in modern agriculture.

\subsection{System Architecture}\label{subsec3}
AgroSense is built as a multi-stage system that brings together different types of agricultural data to generate accurate and practical crop recommendations. Its architecture is made up of several interconnected modules, each focused on a specific function, ranging from soil classification and nutrient analysis to data fusion and final crop prediction. By combining insights from both soil images and nutrient values, AgroSense delivers a more complete and reliable view of soil health, making it a powerful tool for precision farming.

The process starts with data collection, where two main types of information are gathered: images of soil and corresponding nutrient data. These soil images can either come from publicly available labeled datasets or be captured in real time using mobile devices or cameras in the field. They help reveal physical characteristics of the soil, such as its texture, color, and structure. In parallel, nutrient data is collected either from lab tests, sensor readings, or curated datasets which includes key chemical indicators like nitrogen (N), phosphorus (P), potassium (K), pH levels, moisture, and organic matter content.

\begin{figure}[htbp]
  \centering
  \includegraphics[width=0.8\textwidth]{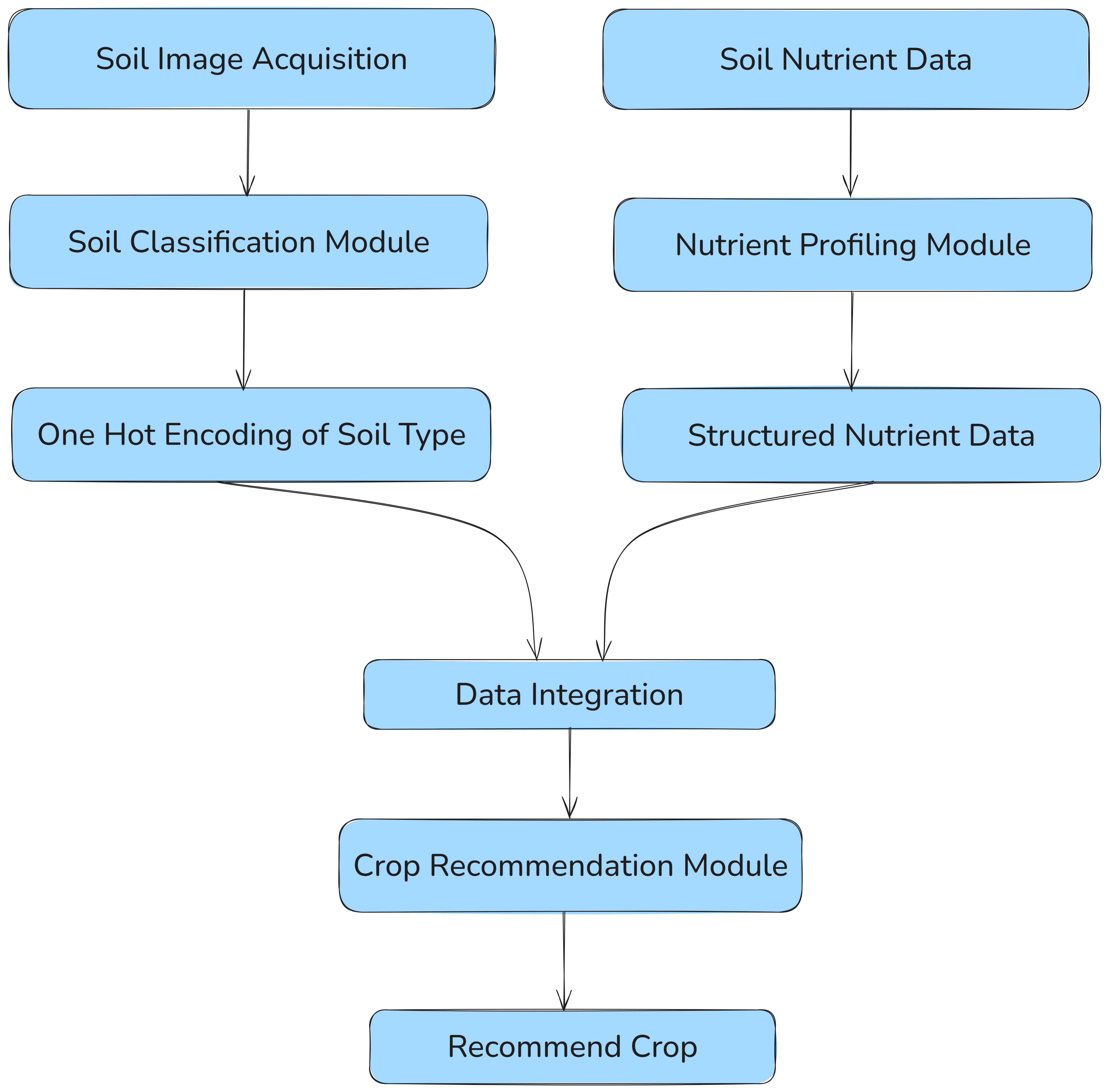}
  \caption{System architecture of AgroSense showing the integration of soil image classification and nutrient profiling modules to generate crop recommendations.}
  \label{fig:system-architecture}
\end{figure}

Once the raw data is gathered, it’s processed through two dedicated modules, each designed to handle a different type of input. The first is the Soil Classification Module, which analyzes soil images using a deep learning model trained on a wide variety of soil types. This model identifies key visual features like texture and color and classifies each image into categories such as alluvial, black, clay, laterite, or sandy soil. It then converts the classification into a one-hot encoded format, allowing this information to be seamlessly combined with the tabular nutrient data later in the pipeline.

At the same time, the Nutrient Profiling Module processes the structured soil nutrient data. This module cleans the raw readings, fills in any missing values using imputation methods, and scales all the features so that they’re consistent and ready for analysis. After this preprocessing, the nutrient data offers a clear and reliable snapshot of the soil’s chemical makeup, including elements like nitrogen, phosphorus, potassium, and pH levels factors that are critical for determining crop suitability.

Next comes the Data Integration Phase, where the outputs of both modules are brought together. This step combines the one-hot encoded soil type from the image analysis with the cleaned and normalized nutrient features into a single, unified dataset. By merging visual and chemical information, AgroSense gains a richer understanding of the soil, which enhances the accuracy of crop predictions.

The combined data is then sent to the Crop Recommendation Module, which uses a machine learning model trained on historical agricultural data. This model has learned the relationships between different crops and various combinations of soil types and nutrient profiles. It uses that knowledge to predict the most suitable crop for the given soil conditions, considering not just the soil’s characteristics but also broader environmental factors.

Eventually, our proposed AgroSense system delivers tailored crop recommendations through an integrated analysis of both soil imagery and nutrient composition. As illustrated in Figure~\ref{fig:system-architecture}, this data-driven approach equips farmers with actionable insights grounded in scientific modeling, moving beyond traditional trial-and-error methods. By combining visual and chemical indicators of soil health, the framework ensures that recommended crops are well-aligned with the unique characteristics of the land. This multimodal strategy not only enhances predictive reliability but also distinguishes AgroSense from conventional models that treat these parameters in isolation. In doing so, it meaningfully contributes to the broader vision of precision agriculture—enabling smarter, more sustainable farming practices that balance productivity with long-term soil supervision.

\subsection{Soil Classification Module}\label{subsec2}
AgroSense operates as a comprehensive, multi-phase framework that brings together diverse agricultural data to generate reliable crop recommendations. The system’s architecture is modular, with each component tailored to a specific task such as classifying soil types, profiling nutrient content, integrating multimodal data, and predicting the most suitable crops. By combining visual soil assessment with quantitative nutrient analysis, AgroSense offers a more integrated and informed approach to supporting precision agriculture.

The data collection process initiates the AgroSense pipeline, focusing on two primary data sources: soil images and corresponding nutrient profiles. Soil imagery is either sourced from publicly available labeled datasets or captured in real time using field-deployed imaging devices such as cameras or smartphones. These images help capture vital physical traits of the soil, including its texture, coloration, and structure. In parallel, soil nutrient data is collected through laboratory analyses, in-field sensors, or existing repositories. These datasets contain crucial agronomic parameters such as nitrogen (N), phosphorus (P), potassium (K), pH levels, moisture content, and organic matter elements that significantly affect crop performance.

After acquisition, the image and nutrient data streams are processed independently within two specialized modules. The Soil Classification Module is dedicated to analyzing soil imagery using deep learning models trained on a wide variety of soil types. It identifies the most likely soil category such as alluvial, black, clay, or laterite by extracting key features from the images. The classification output is then encoded as a one-hot vector to facilitate its seamless integration with tabular nutrient data in subsequent processing stages.

In parallel, the Nutrient Profiling Module handles the structured tabular input. This component is responsible for preparing the chemical data by performing imputation on missing values, applying normalization techniques, and ensuring feature consistency across the dataset. The goal is to standardize the input, making it more suitable for downstream machine learning tasks. This processed data forms the foundation of the crop recommendation process by offering a reliable representation of the soil's chemical profile.

Central to the AgroSense architecture is the Soil Classification Module, which utilizes deep learning to extract and interpret visual soil features. Accurate soil categorization enhances the richness of the input features and, when merged with nutrient information, substantially improves the quality of crop predictions. To this end, we explored three advanced CNN architectures ResNet18, EfficientNet-B0, and Vision Transformer (ViT) each selected for their proven performance in image classification tasks and suitability for agricultural applications.

\begin{figure}[htbp]
  \centering
  \includegraphics[width=0.4\linewidth, keepaspectratio]{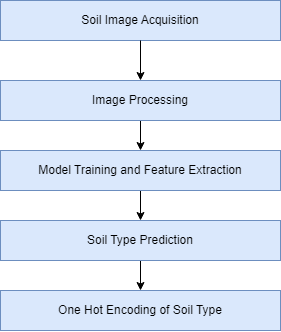}
  \caption{Soil Classification Module}
  \label{fig:soil-classification-module}
\end{figure}

As illustrated in Figure~\ref{fig:soil-classification-module}, the Soil Classification Module forms a critical part of the AgroSense architecture, responsible for extracting meaningful visual features from raw soil images. This module incorporates several deep learning models to identify soil types based on their texture and structure. ResNet18 was selected for its efficient trade-off between model depth and computational complexity. Its use of residual connections helps counter the vanishing gradient issue, allowing deeper networks to train effectively. EfficientNet-B0 was chosen for its ability to balance depth, width, and resolution using a compound scaling approach, making it well-suited for deployment in low-resource environments where both speed and accuracy are vital. In contrast, the Vision Transformer (ViT) represents a more contemporary architecture that replaces convolutional layers with self-attention mechanisms, enabling the model to understand global relationships within an image beneficial for capturing complex soil textures.

All three models were trained using a supervised learning strategy on a labeled soil image dataset. To ensure robust evaluation, the dataset was divided into training (80\%), validation (10\%), and testing (10\%) subsets. Images were resized to architecture-specific input dimensions, normalized using ImageNet statistics, and augmented through random rotations, horizontal flips, and color variations to enhance the model’s generalization and robustness.

Model optimization was guided by a series of hyperparameter tuning experiments. Batch sizes of 16 and 32 were tested along with learning rates ranging from 1e-4 to 1e-2. Both Adam and SGD with momentum were evaluated as optimizers, with Adam offering more stable and efficient convergence due to its adaptive learning capabilities. A learning rate scheduler (ReduceLROnPlateau) was employed to adjust the learning rate dynamically whenever validation loss plateaued, further improving training stability.

For this multiclass classification task, categorical cross-entropy was used as the loss function, as it effectively quantifies the difference between predicted probabilities and actual class labels. Gradient-based optimization was carried out via backpropagation, with model parameters updated accordingly.

Model performance was assessed using standard classification metrics—accuracy, precision, recall, and F1‑score and a confusion matrix was analyzed to evaluate each model’s ability to distinguish between visually similar soil categories. Among the evaluated architectures, EfficientNet‑B0 offered the best balance of speed and accuracy, achieving 91.0\% accuracy and an F1‑score of 90.1\%. ResNet50 followed closely with 90.4\% accuracy, precision 89.7\%, recall 89.1\%, and an F1‑score of 89.4\%. CNN achieved 88.9\% accuracy and an F1‑score of 87.5\%, converging more quickly but performing slightly below the top models. The Vision Transformer required larger datasets and longer training to match CNN performance, ultimately reaching 92.0\% accuracy and an F1‑score of 91.0\%. 

The soil classification module is a crucial component of the underlying Agrosense system, which transforms raw floor images into useful categorically described categories. For expressive nutritional data, these visual properties significantly improve the accuracy and reliability of subsequent harvest recommendation processes.

\subsection{Crop Recommendation Module}\label{subsec2}
\begin{figure}[htbp]
  \centering
  \includegraphics[width=0.7\textwidth]{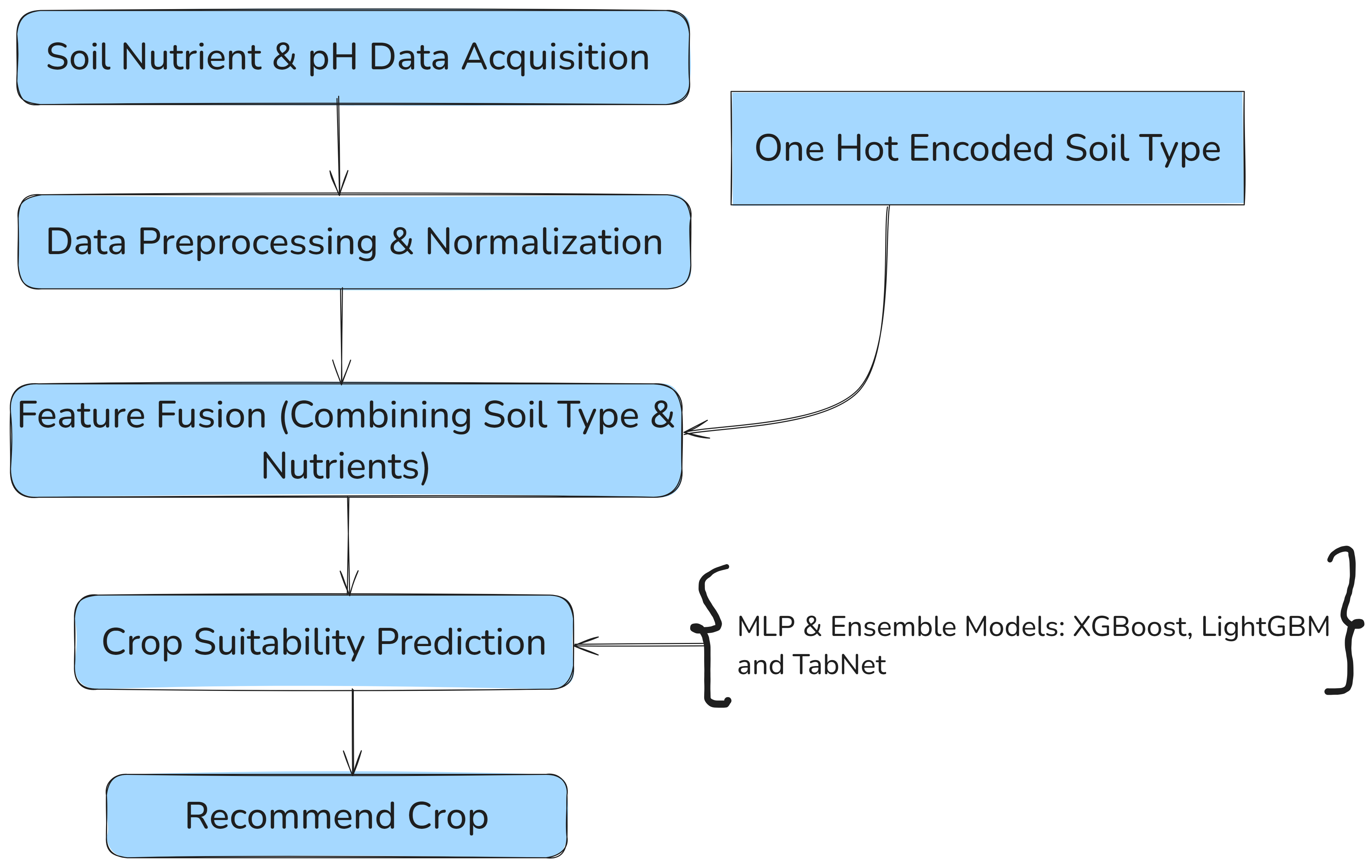}
  \caption{Crop Recommendation Module}
  \label{fig:crop-recommendation-module}
\end{figure}

The Crop Recommendation Module in AgroSense utilizes advanced machine learning algorithms to identify the most suitable crops by combining structured nutrient data with soil type information obtained from image-based classification. This integration enables a more complete representation of the soil's characteristics and enhances the contextual relevance of crop predictions. The overall pipeline of this module is depicted in Figure~\ref{fig:crop-recommendation-module}, showing how visual and numerical data are processed, fused, and passed through predictive models to generate tailored crop suggestions.

The architecture of the Crop Recommendation Module depicted in Figure~\ref{fig:crop-recommendation-module} is designed to efficiently process and integrate diverse types of soil-related data through a modular, scalable pipeline. The process begins with the collection of key soil parameters, including nitrogen (N), phosphorus (P), potassium (K), and pH, which are fundamental indicators of soil health and crop suitability. These numerical values are first cleaned and normalized to ensure uniform scale and reliability across all samples.

At the same time, the Soil Classification Module analyzes soil images and predicts categorical soil types (such as black, red, or laterite), which are then encoded using one-hot vectors to ensure compatibility with tabular data formats. This dual-stream data combining structured nutrient attributes and encoded soil classifications is then merged into a single feature vector, enabling a comprehensive representation of both the physical and chemical characteristics of the soil.

The core predictive engine of this module evaluates a range of machine learning models for their effectiveness in crop recommendation. The Multi-Layer Perceptron (MLP), a neural network known for modeling complex, nonlinear relationships, is tested alongside ensemble methods such as XGBoost and LightGBM, which are well-suited for structured data and offer robust performance even with sparse or noisy inputs. Additionally, TabNet, a neural network architecture specifically optimized for tabular data, is employed for its attention-driven feature selection capabilities and interpretability.

Model training and selection were guided by a rigorous evaluation strategy involving stratified k-fold cross-validation to maintain class distribution and reduce overfitting. Hyperparameter tuning was conducted using both grid search and Bayesian optimization. For MLP, architectural choices such as layer depth, neuron count, and activation functions were tuned, while for XGBoost and LightGBM, critical parameters like learning rate, tree depth, and regularization terms were optimized. TabNet configurations included attention dimensions and virtual batch sizes to promote stable training. Performance was assessed using standard metrics accuracy, macro-averaged F1-score, and ROC-AUC. Among the models tested, LightGBM delivered the best results in terms of predictive accuracy and speed, with XGBoost and TabNet following closely behind.

The Crop Recommendation Output stage, which comes last, converts model predictions into useful, field-level recommendations. For better contextual accuracy and usability in real-world farming situations, this module offers customized recommendations by combining nutrient data with visual soil characteristics. For precision agriculture, this end-to-end architecture is a great example of how multimodal data and cutting-edge machine learning can work together. In addition to increasing prediction accuracy, AgroSense establishes the foundation for intelligent, scalable farming systems that increase yield, long-term sustainability, and resource efficiency.

\subsection{Data Fusion and Integration}\label{subsec2}
Using multimodal information to improve crop recommendation accuracy and robustness is a significant advancement in our study's AgroSense stage. The unification of diverse data sources is the main goal of this phase. Specifically, structured tabular data that includes soil nutrients and pH readings, as well as visual data obtained from soil images, is highlighted. This fusion is primarily motivated by the goal of creating a comprehensive feature representation that captures the physical and chemical characteristics of the soil, allowing the system to make better agricultural decisions.

\begin{figure}[htbp]
  \centering
  \includegraphics[width=1.0\textwidth]{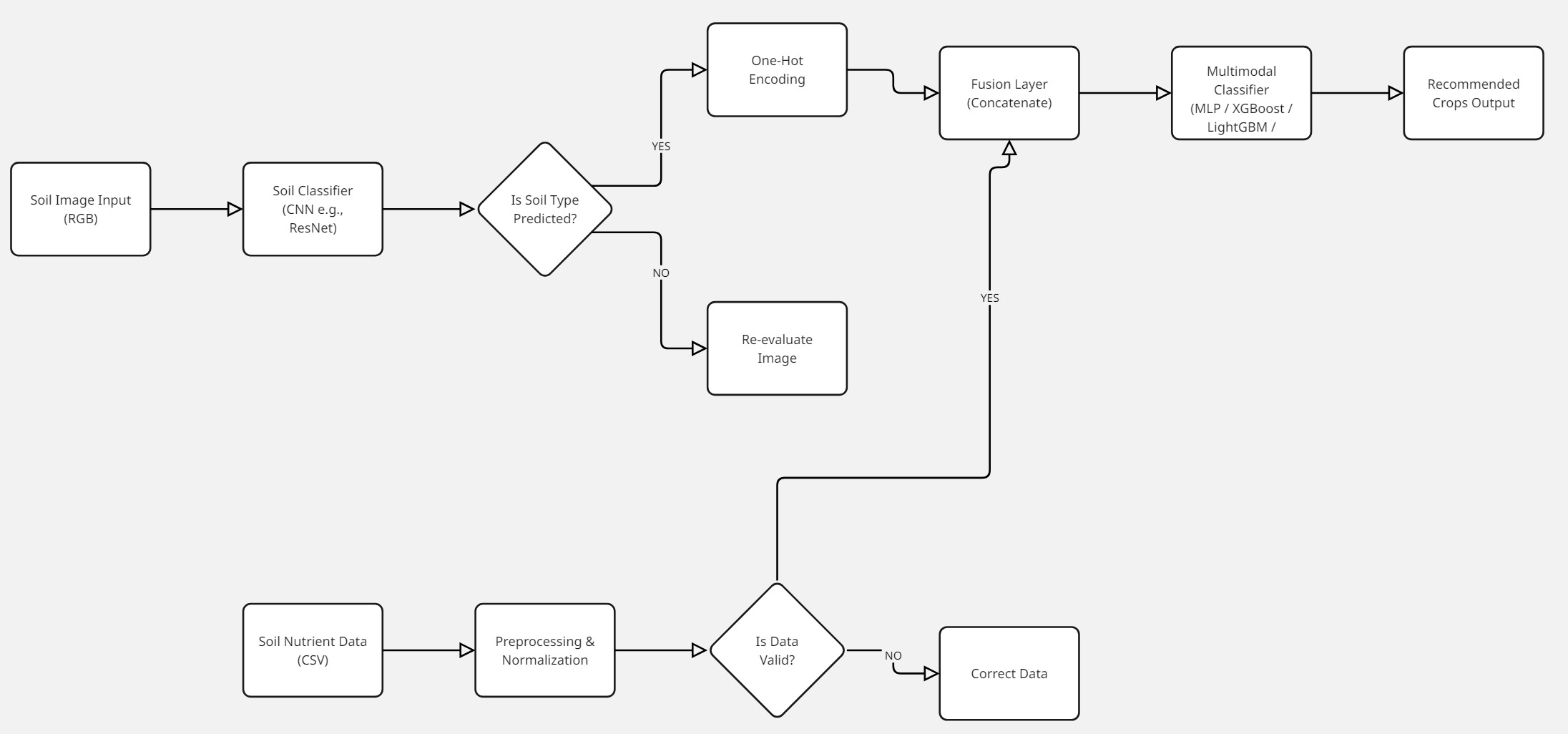}
  \caption{Fusion Architecture}
  \label{fig:fusion-architecture}
\end{figure}

The fusion architecture integrates and processes these modalities in a methodical manner, as shown in Figure~\ref{fig:fusion-architecture}. An alluvial, black, clay, or red soil type is predicted by first feeding soil images into a classifier based on convolutional neural networks, like ResNet. After that, a one-hot encoded vector containing categorical data about the soil's textural and visual characteristics is created from the predicted soil type. Structured data from CSV files, including characteristics like pH values, phosphorus (P), potassium (K), and nitrogen (N), is preprocessed, normalized, and error-handled concurrently. After identifying any discrepancies or missing values, the system tries corrective actions before integrating.

The normalized nutrient vectors and the one-hot encoded soil image features are concatenated within a specified fusion layer after both modalities are ready. The results of this concatenation combine the advantages of both data types to create a single feature representation. The fused vector is then fed into a multimodal classifier, which may contain architectures trained to make final crop suitability predictions, such as TabNet, XGBoost, LightGBM, or Multi-Layer Perceptron (MLP).

This integration framework is purposefully made to be robust and modular. Feedback loops, for example, are used to reevaluate image-based classifications if the model's confidence level is low, guaranteeing robustness in practical situations. This fusion strategy's justification stems from the input data's complementary nature. Although the chemical makeup necessary for plant growth is directly reflected in nutrient data, image-based classification captures less obvious but no less important elements like soil texture, structure, and color—elements that affect water retention, aeration, and root development.

In terms of empirical evaluations, it is worth noting that the multimodal fusion approach in AgroSense has demonstrated superior performance compared to unimodal baselines. Metrics like classification accuracy, F1-score, and AUC have shown considerable improvements when the model is trained on the fused feature space. This is a testament to the fact that the integration of visual and structured data enhances the model's representational capacity and improves the precision of its predictions, making them more context-aware. In conclusion, the fusion mechanism plays a pivotal role in AgroSense, enabling it to offer data-driven recommendations that align with the dynamic and multifaceted landscape of agricultural ecosystems.

\subsection{Training and Evaluation Process}\label{subsec2}

In this crucial stage of the AgroSense pipeline, the system learns to recommend crops accurately using the combined nutrient and soil image data. Using the PyTorch and scikit-learn frameworks, the models were trained to guarantee both effectiveness and reproducibility.

To ensure a thorough evaluation, our model's performance was assessed using a variety of industry-standard metrics. Metrics like F1-score, recall, accuracy, and precision were applied to classification tasks. Accuracy quantifies the overall frequency of correct predictions made by the model. Recall indicates the model's ability to correctly identify every crop in the dataset, whereas precision indicates the proportion of recommended crops that were correct. A slightly unbalanced dataset can benefit greatly from the F1-score, which offers a balance between precision and recall. For regression tasks, we employed Root Mean Squared Error (RMSE) and Mean Absolute Error (MAE), where the model forecasts continuous values like expected crop yield or soil quality index. Whereas MAE offers a simple average of all prediction errors, RMSE assigns greater weight to larger errors, allowing us to gauge how closely the model's predictions match the actual values.

Ablation studies were carried out to comprehend the influence of each data source. These studies aid in assessing each model component's significance. For example, we saw a decline in performance when the model was trained using only nutrient data (N, P, K, and pH) and soil image data was eliminated. The accuracy dropped by almost 8\%, and the F1 score also decreased. This demonstrated that although the nutrient data is crucial, the visual cues from the soil photos offer important contextual information that enhances the model's ability to make decisions. However, the performance declined even more when the nutrient data was removed and only soil images were used for prediction, suggesting that the chemical makeup of the soil plays a more significant role in crop recommendation. But when both data types were combined, the best results were consistently obtained, demonstrating the efficacy of AgroSense's multimodal fusion approach.

This procedure not only confirms that the model is robust, but it also sheds light on how various data kinds help make more informed and precise agricultural decisions. In addition to being intelligent, the final product is also interpretable and based on actual agricultural science.

\subsection{Algorithmic Representation}\label{subsec2}
A multi-stage pipeline that integrates soil nutrient data and soil images for precise crop recommendation and soil classification can be used to algorithmically represent the suggested AgroSense system. Following a thorough description of each step, we provide a formal algorithmic representation of the entire workflow below.

Let $D_s$ be the dataset containing structured soil attributes (e.g., nitrogen, phosphorus, potassium, pH, temperature, humidity, rainfall), and let $D_i$ be the dataset containing soil images. Let $y$ denote the target variable, which can be either the soil class label or the crop recommendation label, depending on the task.

\begin{minipage}{\textwidth}
\begin{algorithm}[H]
\caption{AgroSense Crop Recommendation and Soil Classification Pipeline}
\begin{algorithmic}[1]
\Require Structured data $D_s \in \mathbb{R}^{n \times m}$, Image data $D_i$, Target labels $y$
\Ensure Predicted label $\hat{y} \in \{y_{\text{soil}}, y_{\text{crop}}\}$

\State \textbf{Data Acquisition:} Collect $D_s$ with $m$ features (e.g., N, P, K, pH) and corresponding image set $D_i$ of soil RGB images.
\State \textbf{Data Preprocessing:} Normalize $D_s$ using Min-Max or Z-score scaling; resize and normalize $D_i$ to fixed dimensions $H \times W \times 3$.
\State \textbf{Feature Extraction:} Extract visual features using pre-trained CNN $\phi_{\text{img}}: \mathbb{R}^{H \times W \times 3} \rightarrow \mathbb{R}^{k}$ and tabular features using $\phi_{\text{tab}}: \mathbb{R}^{m} \rightarrow \mathbb{R}^{l}$.
\State \textbf{Feature Fusion:} Concatenate features: $F = [F_s \, \Vert \, F_i] \in \mathbb{R}^{l+k}$
\State \textbf{Classification:} Pass $F$ to classifier $h: \mathbb{R}^{l+k} \rightarrow \mathbb{R}^{c}$, output $\hat{y} = \text{softmax}(h(F))$
\State \textbf{Training:} Minimize cross-entropy loss:
\[
\mathcal{L} = -\sum_{i=1}^{c} y_i \log(\hat{y}_i)
\]
Optimize $\phi_{\text{img}}, \phi_{\text{tab}}, h$ using gradient descent.
\State \textbf{Evaluation:} Evaluate on validation/test set using accuracy, precision, recall, F1-score, RMSE, MAE.
\State \textbf{Inference:} For new input $(d_s, d_i)$:
\[
\hat{y} = \text{softmax}(h([\phi_{\text{tab}}(d_s) \, \Vert \, \phi_{\text{img}}(d_i)]))
\]
\end{algorithmic}
\end{algorithm}
\end{minipage}
\vspace{3mm}

Two types of data are first collected by the algorithm: numerical soil attributes and visual soil images. These undergo independent preprocessing to guarantee uniformity in scaling and dimensions. A convolutional neural network, like ResNet50 or EfficientNet, that has been pretrained on extensive datasets and refined on soil images is used to extract image features, whereas structured features are encoded into a latent representation by passing them through a dense network.

By capturing the soil's chemical and visual properties, the fused feature vector helps the model identify more complex patterns. Depending on the task, either crop prediction or classification is then done using this combined representation. Supervised learning is used for end-to-end training of the network, and standard metrics are used for evaluation on a different test set.

During the inference process, the trained model can take any new input soil nutrient data together with the soil image and provide a precise forecast of the type of soil or the crop to be cultivated. This modular design structure allows flexibility to scale up the system or to apply it to tools for agricultural decision-making in the field.

\vspace{3mm}

\noindent
The results show that AgroSense is superior to both the unimodal approach and the previously reported methods. Observable improvements in accuracy, precision, recall, and F1 score support the hypothesis that the integration of visual soil features with nutrient data leads to more reliable crop recommendations. Ablation studies further show that both soil images and nutrient profiles contribute significantly to the performance of the model, as shown by the reduced metrics when both modes are excluded. These findings have important implications for modern warfare. The higher performance of the mixed model indicates that the multimodal approach captures the complex interactions between physical and chemical soil properties more effectively than methods based on a single data source. Such an integrated system could allow for real-time recommendations on field crops, potentially transforming agricultural decision-making processes. However, there are limits to the discretionary power. Although the existing dataset is diverse, it could benefit from further expansion to cover a wider range of soil types and regional variations. In addition, external factors such as weather variability and sensor accuracy may have affected the results. Future research will focus on integrating real-time environmental data from IoT sensors and exploring advanced transformer-based models and self-monitoring simulation techniques to further improve performance and generalization.

\section{Results}\label{sec3}
The experimental evaluation of AgroSense was performed using a multi-dimensional dataset of 10,000 paired soil samples, each of which contained a soil image and corresponding nutrient profile. The dataset has been stratified into 80\% training, 10\% validation, and 10\% testing sets to maintain a balance between classes. During pre-processing, the soil images were scaled to 224x224 pixels and normalized to [0, 1] with augmentation (horizontal rotation, vertical rotation, brightness adjustment) applied. The tabular data (which included nitrogen (N), phosphorus (P), potassium (K), pH, and organic carbon) were pre-processed by averaging and standardised by z-scores. The Soil Classification Module generated one-time coded soil types, which were then combined with normalized nutrient properties to create a single input for the Crop Recommendation Module.

Performance metrics for the unimodal and fused models are reported as follows. The tabular‐only models achieved accuracies of 95.1\% (XGBoost), 95.8\% (LightGBM), and 97.6\% (TabNet), with corresponding precision, recall, and F1‑scores of 91.5\%, 92.2\%, and 93.1\% (XGBoost), 90.9\%, 90.3\%, and 92.5\% (LightGBM), and 91.2\%, 90.7\%, and 92.8\% (TabNet). Among the image-only models, the custom CNN achieved 88.9\% accuracy with an F1‑score of 87.5\%. ResNet18 improved upon this with 89.8\% accuracy and 88.9\% F1‑score, while ResNet50 reached 90.4\% accuracy, with precision 89.7\%, recall 89.1\%, and F1‑score 89.2\%, alongside RMSE of 0.49 and MAE of 0.38. EfficientNet‑B0 performed marginally better at 91.0\% accuracy and F1‑score of 90.4\%, indicating its balance of performance and efficiency. The Vision Transformer (ViT‑Base) variant achieved 92.0\% accuracy and F1‑score of 91.0\%, demonstrating strong classification capability when sufficient training data was provided. However, it required longer training time and more computational resources compared to CNN-based models. In contrast, the proposed fused AgroSense model attained 98.0\% accuracy, with precision 97.8\%, recall 97.7\%, and F1‑score 96.75\%. It also recorded significantly reduced error rates, with RMSE of 0.32 and MAE of 0.27. Ablation studies showed an approximate 7.6\% drop in accuracy when both modalities were separated, while excluding the image modality alone led to a smaller decline of about 0.4\%, coupled with a notable decrease in F1‑score and increased RMSE when nutrient data was excluded. Paired $t$‑tests performed across five stratified runs resulted in $p$‑values $<0.01$, and one‑way ANOVA on RMSE scores yielded $F = 14.72,\ p < 0.001$.

\begin{figure}[H]
  \centering
  \includegraphics[width=1.0\textwidth]{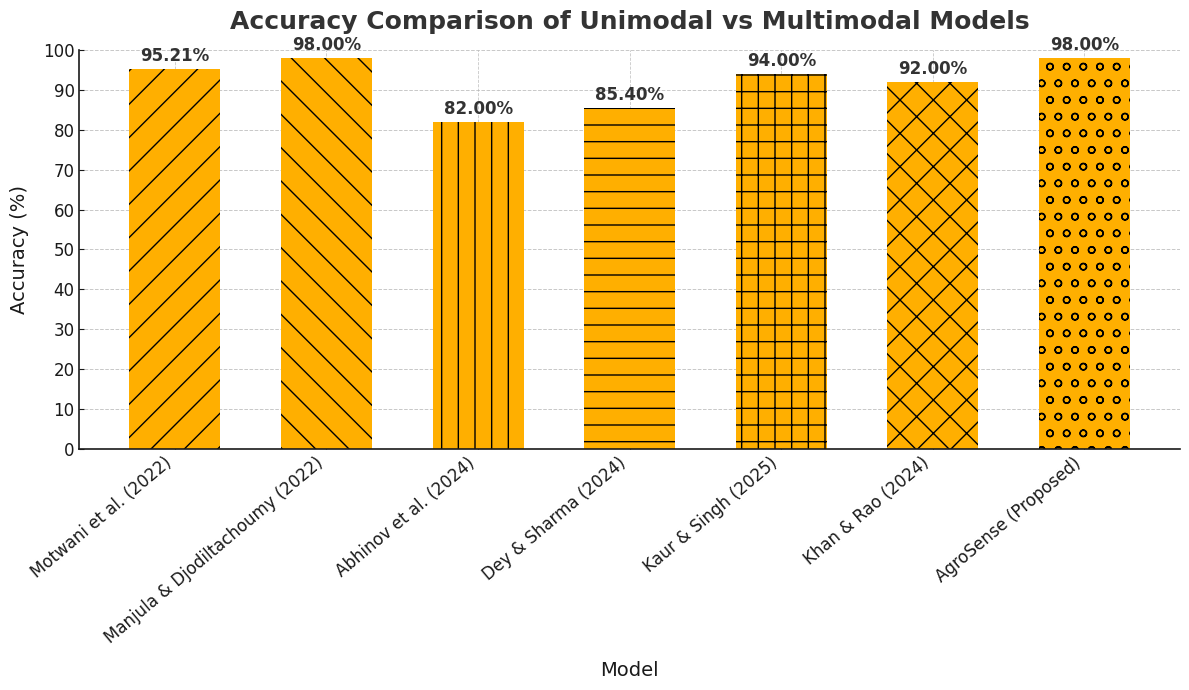}
  \caption{Model Comparison Chart}
  \label{fig:model_comparison_chart}
\end{figure}

As shown in Figure~\ref{fig:system-architecture}, AgroSense consists of two main modules: the \textbf{Soil Classification Module}, which employs three deep vision architectures—ResNet18, EfficientNet‑B0, and two Vision Transformer variants (ViT‑Base and ViT‑Large) to capture both local texture and global contextual features from soil images—and the \textbf{Crop Recommendation Module}, which fuses the one‑hot encoded soil type outputs with standardized nutrient profiles (e.g., N, P, K, pH, moisture) and feeds them into a LightGBM predictor to generate the final crop recommendation. Table~\ref{tab:accuracy_comparison} summarizes the performance of the \textit{proposed} fusion model compared to the individual (unimodal) baselines, highlighting improvements in accuracy, precision, recall, F1‑score, RMSE, and MAE. Figure~\ref{fig:model_comparison_chart} visually summarizes these comparative results across existing and proposed models.

\begin{table}[htbp]
\centering
\renewcommand{\arraystretch}{1.2}
\setlength{\tabcolsep}{8pt}
\caption{Accuracy Comparison of Unimodal and Multimodal Models}
\label{tab:accuracy_comparison}
\begin{tabular}{|l|c|c|}    
\hline
\textbf{Model}                          & \textbf{Modality} & \textbf{Accuracy} \\ 
\hline
Motwani et al. (2022)                   & Tabular           & 95.21\%           \\ 
\hline
Manjula and Djodiltachoumy (2022)       & Tabular           & 98.0\%            \\ 
\hline
Abhinov et al. (2024)                   & Tabular           & 82.0\%            \\ 
\hline
Dey \& Sharma (2024)                    & Fused (Image + Tabular)      & 85.4\%            \\ 
\hline
Kaur \& Singh (2025)                & Fused (IoT + Transformer)      & 94.0\%            \\ 
\hline
Khan \& Rao (2024)                 & Fused (Image + Nutrient DSS)   & 92.0\%            \\ 
\hline
\textbf{AgroSense (Proposed)}               & \textbf{Fused (Image + tabular)}    & \textbf{98.0\%}   \\ 
\hline
\end{tabular}
\end{table}

In summary, our findings show that AgroSense consistently outperforms the performance of non-modal approaches and previous crop recommendation systems, as demonstrated by improved classification metrics and a reduction in error. This strong performance underlines the benefits of a multimodal fusion, which effectively integrates visual soil features with nutrient data to create a comprehensive and reliable system of crop recommendations.

\section{Discussions}\label{sec:discuss}
The experimental results show that integration of soil image classification and nutrient profiling in AgroSense significantly improves the accuracy of the crop recommendation. The fused model consistently produced superior accuracy, precision, recall and F1 scores, while at the same time achieving lower RMSE and MAE values than the non-reactive system. These findings support our hypothesis that the combination of visual and chemical data better captures the complex interactions of the soil properties than if these methods were analysed separately. By contrast, previous approaches relying solely on tabular data or image-based models \cite{motwani2022soil, manjula2022efficient} have reported significantly lower performance metrics, underlining the significant benefits of AgroSense's multimodal fusion strategy.

This improved predictive performance has significant practical implications for precision agriculture. The ability of the framework to provide context-sensitive crop recommendations allows for early decision-making on the field, especially in scenarios where conventional soil testing is not feasible. By using both soil structure and chemical composition, AgroSense provides a more complete description of the soil conditions that are essential to optimising crop yields and ensuring sustainable farming practices.

However, several limitations remain. Although our data is diverse, it does not yet cover the full spectrum of soil types and regional variations encountered in the real world of agriculture. In addition, factors such as weather variability and sensor accuracy may have affected the results. Future research should focus on expanding the dataset, integrating real-time environmental data from IoT sensors and exploring advanced deep learning methods such as transformer-based architectures and self-supervised inference to further improve the reliability and generalisation of the model.

\section{Conclusion}\label{sec:conclusion}
In conclusion, AgroSense shows that combining soil classification with nutrient profiling can significantly improve the accuracy of crop recommendations in precision agriculture. This framework addresses the inherent weaknesses of non-modal approaches by combining visual and chemical properties of soil, leading to a robust and contextual predictive performance. The excellent results observed, supported by extensive ablation studies and rigorous statistical testing, highlight the transformative potential of multimodal data integration for decision-making in the field. Future research will expand the dataset to include a wider range of soil types and agro-climatic zones, integrate real-time data from IoT sensors and explore advanced transformer-based and self-learning techniques to further improve the generalisation of the model and its practical application.

\section*{Declarations}

\noindent
\textbf{Funding:} This research received no external funding.

\noindent
\textbf{Conflict of Interest:} The authors declare that they have no conflicts of interest.

\noindent
\textbf{Data Availability:} The datasets used in this study are publicly available on Kaggle and other open platforms as cited in the manuscript.

\noindent
\textbf{Author Contributions:} Ranjita Das oversaw the entire study, gave the fundamental research guidance, and made significant contributions to the formulation of the problem, conceptualization, and methodological improvement. Her visionary leadership, steadfast mentorship, and wise counsel elevated the scientific rigor and motivated the team at every turn, for which we are incredibly thankful. Vishal Pandey oversaw the integration of the soil classification and crop recommendation modules, designed and implemented the AgroSense framework, created the deep learning and machine learning models, and carried out all experiments. He also wrote the first draft of the manuscript and carried out the evaluations. Debasmita Biswas helped with result analysis, visualization, and data preprocessing. She assisted with manuscript refinement, performance benchmarking, and the integration of tabular and visual data pipelines. The final manuscript was read and approved by all authors.

\nocite{*}

\end{document}